\documentclass[graybox]{svmult}

\usepackage{mathptmx}       
\usepackage{helvet}         
\usepackage{courier}        
\usepackage{type1cm}        
\usepackage{makeidx}         
\usepackage{graphicx}        
\usepackage{multicol}        
\usepackage[bottom]{footmisc}

\usepackage{amsmath} 
\usepackage{amssymb}  
\usepackage{threeparttable}
\usepackage{subfigure}

\usepackage{marvosym}
\newcommand{\envelope}{(\raisebox{-.5pt}{\scalebox{1.45}{\Letter}})}

\graphicspath{{figs/}}

\newcommand\T{\rule{0pt}{2.6ex}}        
\newcommand\B{\rule[-1.2ex]{0pt}{0pt}} 

\def\Vec#1{\mathbf{#1}}
\newcommand{\bbm}{\begin{bmatrix}}
\newcommand{\ebm}{\end{bmatrix}}

\makeindex             

\begin{document}

\title*{Monocular Visual Teach and Repeat Aided by Local Ground Planarity}
\author{Lee Clement\and Jonathan Kelly \and Timothy D. Barfoot}
\institute{Lee Clement \envelope \and Jonathan Kelly \and Timothy D. Barfoot \at Institute for Aerospace Studies, University of Toronto, Toronto, Canada \\ \email{ \{lee.clement@mail., jkelly@utias., tim.barfoot@\}utoronto.ca } }
%
%
\maketitle

\newcommand{\abstracttext} {
Visual Teach and Repeat (VT\&R) allows an autonomous vehicle to repeat a previously traversed route without a global positioning system.
Existing implementations of VT\&R typically rely on 3D sensors such as stereo cameras for mapping and localization, but many mobile robots are equipped with only 2D monocular vision for tasks such as teleoperated bomb disposal.
While simultaneous localization and mapping (SLAM) algorithms exist that can recover 3D structure and motion from monocular images, the scale ambiguity inherent in these methods complicates the estimation and control of lateral path-tracking error, which is essential for achieving high-accuracy path following.
In this paper, we propose a monocular vision pipeline that enables kilometre-scale route repetition with centimetre-level accuracy by approximating the ground surface near the vehicle as planar (with some uncertainty) and recovering absolute scale from the known position and orientation of the camera relative to the vehicle.
This system provides added value to many existing robots by allowing for high-accuracy autonomous route repetition with a simple software upgrade and no additional sensors.
We validate our system over 4.3 km of autonomous navigation and demonstrate accuracy on par with the conventional stereo pipeline, even in highly non-planar terrain.
}

\abstract*{\abstracttext}

\abstract{\abstracttext}

\section{Introduction}

Visual Teach and Repeat (VT\&R) is an effective tool for autonomously navigating previously traversed paths using only on-board visual sensors.
In an initial \textit{teach pass}, a human operator manually drives an autonomous vehicle along a desired route while the VT\&R system uses imagery from a camera to build a map of the route.
In the subsequent \textit{repeat pass}, the system localizes against the stored map to autonomously repeat the route, sometimes combining map-based localization with visual odometry (VO) to estimate relative motion in cases where map-based localization is temporarily unavailable \cite{Furgale2010}.
VT\&R is well-suited to repetitive navigation tasks where GPS is unavailable or insufficiently accurate, and has found applications in autonomous tramming for mining operations \cite{Marshall2008} and sample return missions \cite{Furgale2010}.

The map representation in a VT\&R system may be purely topological, purely metric, or a mixture of the two (sometimes called topometric).
Purely topological VT\&R \cite{Goedeme2007,Matsumoto1996,Remazeilles2006} uses a network of reference images (keyframes) where the navigation goal is to match the current image to the nearest keyframe using a visual homing procedure.
These systems are restricted to heading-based control, which only loosely bounds lateral path-tracking error.
Purely metric maps are uncommon in VT\&R systems due to the high computational cost of creating globally consistent maps for long routes, but successful applications do exist \cite{Kidono2002,Royer2007}.
Topometric systems \cite{Furgale2010,Marshall2008,Simhon1998,Zhang2009} reap the benefits of both mapping strategies by decoupling map size from path length while still retaining metric information.

\begin{figure}[b]
	\centering
	\includegraphics[width=\textwidth]{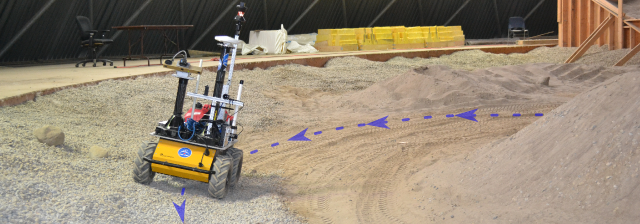}
	\caption{Our field robot during a 140 m autonomous traverse in the UTIAS MarsDome indoor rover testing environment, with the path overlaid for illustration. In order to compare the performance of stereo and monocular VT\&R with the same hardware, we equipped our rover with a stereo camera and used only the left image stream for our monocular traverses.}
	\label{fig:husky_on_path}
\end{figure}

Furgale and Barfoot \cite{Furgale2010} developed the first VT\&R system capable of autonomously repeating multi-kilometre routes in unstructured outdoor terrain using only a stereo camera.
Their system creates a topometric map of metric keyframes connected by 6DOF VO estimates, which are combined via local bundle adjustment into locally consistent metric submaps for localization in the repeat pass.

Furgale and Barfoot's system has been extended to other 3D sensors such as lidar \cite{McManus2013} and RGB-D cameras, but a monocular implementation has not been forthcoming.
While monocular cameras are appealing in terms of size, cost, and simplicity, perhaps the most compelling motivation for using monocular vision for VT\&R is the plethora of existing mobile robots that would benefit from it.
Indeed, vehicles equipped with monocular vision, typically for teleoperation, run the gamut of robotics applications, and in many cases -- search and rescue, mining, construction, and personal assistive robotics, to name a few -- would benefit from accurate autonomous route-repetition, especially if it were achievable with existing sensors.

Several techniques exist for accomplishing online 3D simultaneous localization and mapping (SLAM) with monocular vision, ranging from filter-based approaches \cite{Davison2007,Eade2006} to online batch techniques that make use of local bundle adjustment \cite{Holmes2013,Klein2007,Zhao2010}.
Such algorithms are capable of producing accurate 3D maps, but only up to an unknown scale factor.
This scale ambiguity complicates threshold-based outlier rejection, as well as the estimation and control of lateral path-tracking error during the repeat pass, which are essential for achieving high-accuracy route-following.

In this paper, we extend Furgale and Barfoot's VT\&R system to monocular vision by using the approximately known position and orientation of a camera mounted on a rover to estimate the 3D positions of keypoints near the ground with absolute scale.
Similar techniques have succeeded in computing VO with a monocular camera using both sparse feature tracking \cite{Choi2011,Farraj2013,Zhang2012} and dense image alignment \cite{Lovegrove2011}, but have not considered the problem of map construction.
We show that by treating the ground surface near the vehicle as approximately planar and applying an appropriate uncertainty model, we can generate local metric maps that are accurate enough to achieve centimetre-level accuracy during the repeat pass, even on highly non-planar terrain.
Although the flat-ground assumption is not globally valid, it is sufficient for our purposes since VT\&R uses metric information only locally.

The main contribution of this paper is an extensive comparison of the performance of monocular and stereo VT\&R in a variety of conditions, including an evaluation of their robustness to common failure cases.
To this end, we present experimental results comparing the two systems over 4.3 km of autonomous navigation.
While our results show that both systems achieve similar path-tracking accuracy when functioning normally, the monocular system suffers a reduction in robustness compared to its stereo counterpart in certain conditions.
We argue that, for many applications, the benefit of deploying VT\&R without a potentially costly sensor upgrade far outweighs the associated reduction in robustness.

\section{Monocular Depth Estimation} \label{sec:depth}
We estimate the 3D coordinates of features observed by a camera pointed downward, but not directly at the ground surface, by
approximating the local ground surface near the vehicle as planar and recovering absolute scale from the known position and orientation of the camera relative to the vehicle.
We account for variations in terrain shape by applying an appropriate uncertainty model.
In what follows, $\Vec{z}^i_j$ denotes the 3D coordinates of feature $i$ expressed in coordinate frame $\mathcal{F}_j$.

\subsection{Locally Planar Ground Surfaces} \label{subsec:planarground}
\begin{figure}
    \centering
    \subfigure[Coordinate frames in our monocular depth estimation scheme. The local ground frame $\mathcal{F}_g$ is defined relative to the vehicle frame $\mathcal{F}_v$ and travels with the vehicle.] {
    	\includegraphics[width=0.45\textwidth]{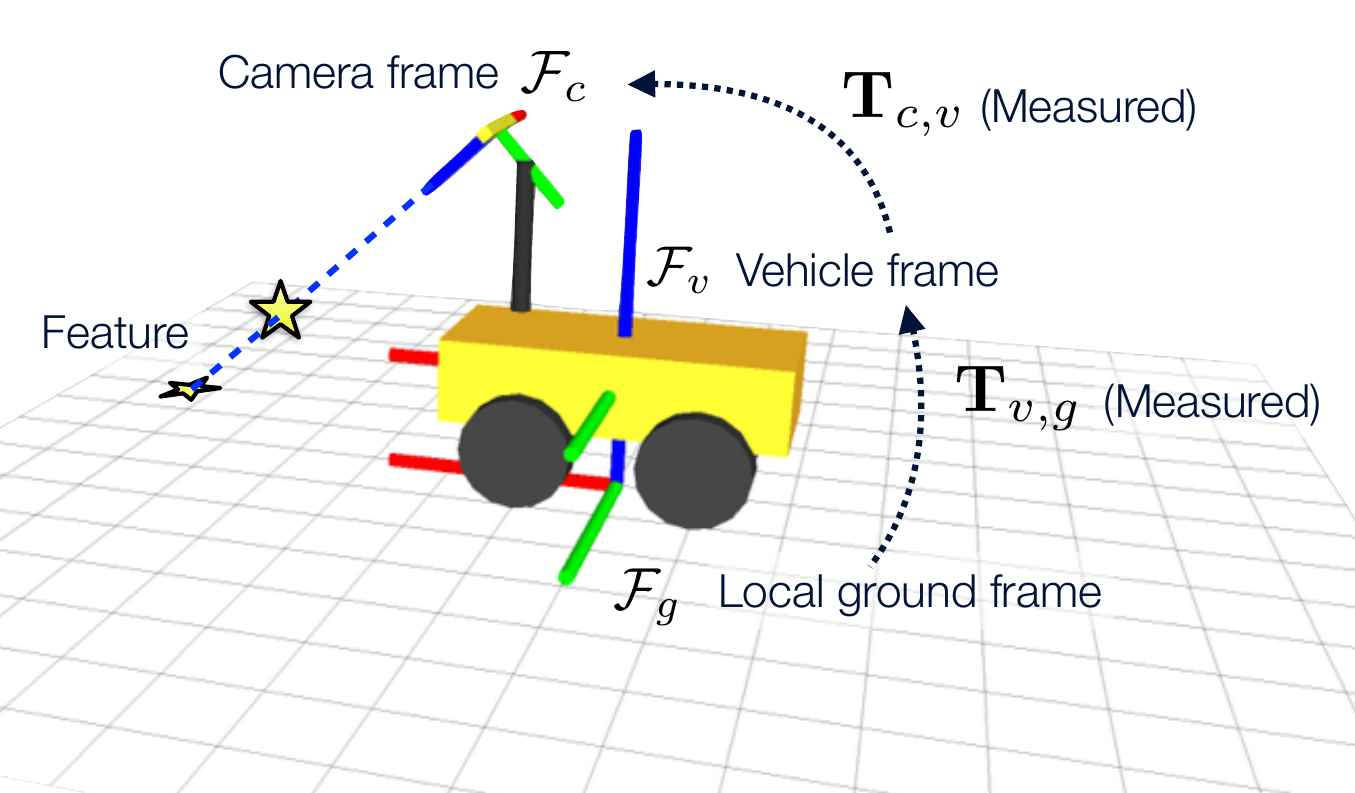}
    	\label{fig:frames}
	}
	~
	\subfigure[Evenly-spaced synthetic image features (top right) and estimated 3D coordinates with $1\sigma$ uncertainty ellipses for the experimental configuration described in Section \ref{sec:experiments}.] {
    	\includegraphics[width=0.45\textwidth]{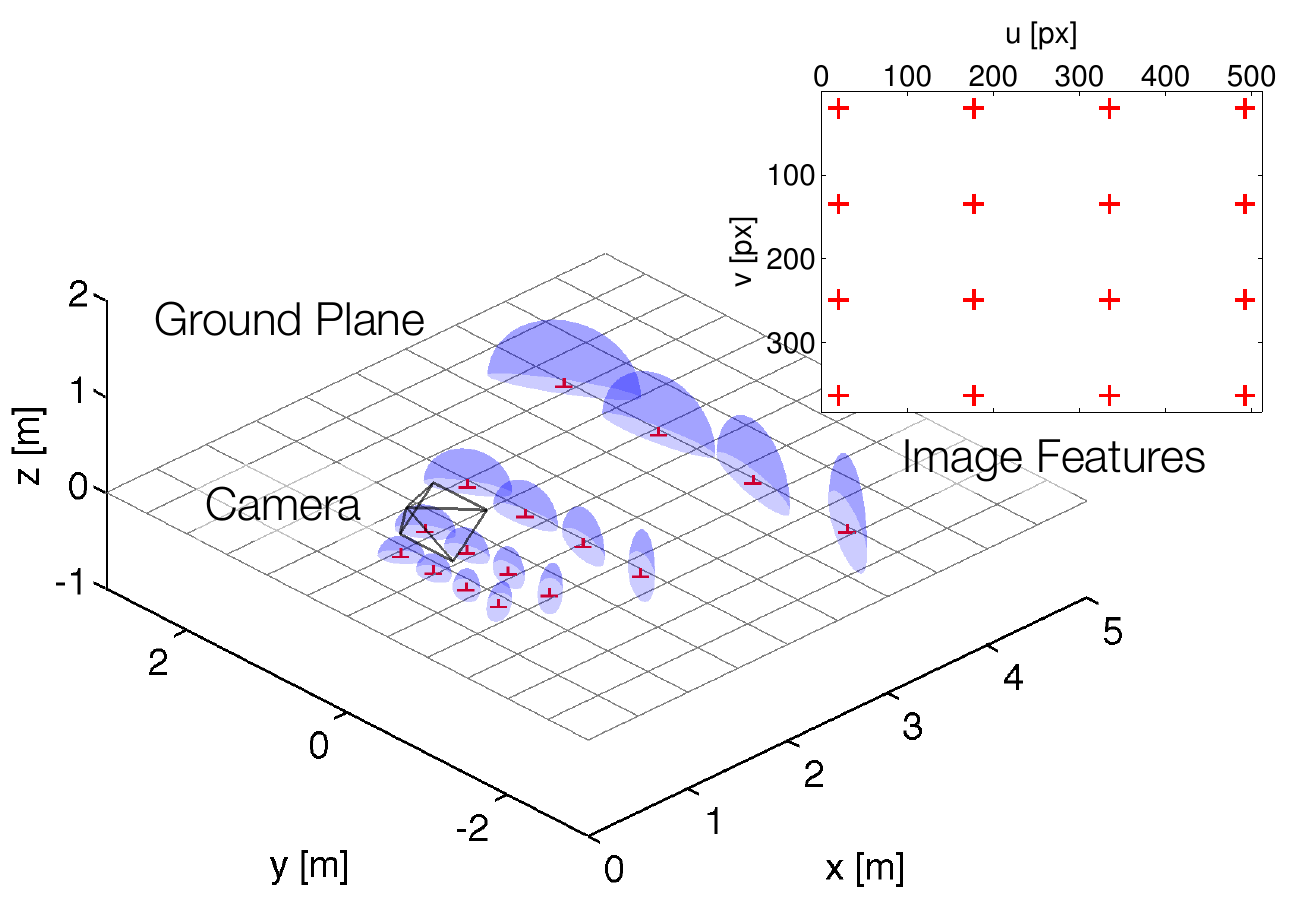}
    	\label{fig:ellipses}
	}
	\caption{Geometry and uncertainty model of our monocular depth estimation scheme.}
\end{figure}

For a monocular camera observing the ground, we can estimate the 3D coordinates of features near the ground by making the following assumptions (see Figure \ref{fig:frames}):
\begin{enumerate}
    \item all features of interest lie in the $xy$-plane of a local ground frame $\mathcal{F}_g$ defined such that its $z$-axis is normal to the ground and always intersects the origin of the vehicle coordinate frame $\mathcal{F}_v$ (for a ground vehicle, this is the vehicle's footprint);
    \item the transformation $\Vec{T}_{c,v}\in\text{SE(3)}$ from $\mathcal{F}_v$ to the camera-centric coordinate frame $\mathcal{F}_c$ is known; and
    \item the transformation $\Vec{T}_{v,g}\in\text{SE(3)}$ from $\mathcal{F}_g$ to $\mathcal{F}_v$ is known.
\end{enumerate}

Assuming that incoming images have been de-warped and rectified in a pre-processing step, we can model the camera as an ideal pinhole camera with calibrated camera matrix $\Vec{K}$
such that the image coordinates $\Vec{y}^i$ of $\Vec{z}^i_c$ are given by
\begin{equation} \label{eq:intrinsics}
\Vec{y}^i := \bbm u^i&v^i&1 \ebm^T = \Vec{K}\Vec{p}^i \text{ ,}
\end{equation}
where
\begin{equation} \label{eq:projection}
\Vec{p}^i := \bbm p^i_x&p^i_y&1 \ebm^T = \dfrac{1}{z^i_c} \bbm x^i_c&y^i_c&z^i_c \ebm^T
\end{equation}
represents the (unitless) normalized coordinates of $\Vec{z}^i_c$ on the image plane.
Note that although $u^i, v^i$ represent pixel coordinates, they are not necessarily integer-valued.

By assumption 1, $z^i_g = 0, \forall i$, so we can write
\begin{equation} \label{eq:featuresonground}
\Vec{z}^i_c := \bbm x^i_c&y^i_c&z^i_c&1 \ebm^T = \Vec{T}_{c,g} \bbm x^i_g&y^i_g&0&1 \ebm^T \text{ ,}
\end{equation}
where $\Vec{T}_{c,g} = \Vec{T}_{c,v}\Vec{T}_{v,g}$.
We can therefore obtain the feature depth $z^i_c$ as a function of $\Vec{p}^i$ by substituting $x^i_c = z^i_cp^i_x$ and $y^i_c = z^i_cp^i_y$ according to Equation \eqref{eq:projection}, and solving the third component of Equation \eqref{eq:featuresonground} for $z^i_c$, yielding
\begin{equation} \label{eq:depth}
z^i_c = \frac{k_1}{k_2 + k_3p^i_x + k_4p^i_y} \text{ ,}
\end{equation}
where, using $T_{mn}$ as shorthand for the $m$th row and $n$th column of $\Vec{T}_{c,g}$,
\begin{align*}
    k_1 &= T_{11}\left(T_{22}T_{34} - T_{24}T_{32}\right) & k_2 &= T_{11}T_{22} - T_{12}T_{21} \\ 
    	&+ T_{12}\left(T_{24}T_{31} - T_{21}T_{34}\right) & k_3 &= T_{21}T_{32} - T_{22}T_{31} \\ 
    	&+ T_{14}\left(T_{21}T_{32} - T_{22}T_{31}\right) & k_4 &= T_{12}T_{31} - T_{11}T_{32} \text{ .}
\end{align*}

Finally, using Equations \eqref{eq:intrinsics} and \eqref{eq:projection} with $z^i_c$ as in Equation \eqref{eq:depth},
we can express the Cartesian coordinates of $\Vec{z}^i_c$ in terms of $\Vec{y}^i$ as
\begin{equation} \label{eq:backprojection}
\Vec{z}^i_c =  z^i_c\Vec{K}^{-1}\Vec{y}^i \text{ .}
\end{equation}

\subsection{Uncertainty Considerations}
\label{sec:uncertainty}

A crucial component of enabling monocular VT\&R using this depth estimation scheme is an appropriate model of the uncertainty in each observation $\Vec{z}^i_c$. 
We consider two important factors: uncertainty in image coordinates $\Vec{y}^i$, and uncertainty in ground shape far from the vehicle.
In early experiments, we found that image coordinate uncertainty alone did not permit reliable feature tracking since there was little overlap in 3D feature coordinate estimates across multiple frames.

We model feature coordinates in image space as Gaussian distributions centred on $\Vec{y}^i$ with covariance $\Vec{R}_{\Vec{y}^i} := \text{diag}\{(\sigma^i_u)^2, (\sigma^i_v)^2\}$.
We use SURF features \cite{Bay2008} in our system and determine $\sigma^i_u, \sigma^i_v$ from the image pyramid level at which each feature is detected.
To incorporate uncertainty in ground shape far from the vehicle, we represent the ground-to-vehicle transformation as a Gaussian distribution on SE(3) with mean $\Vec{T}_{v,g}$ and covariance
$\Vec{R}_{\Vec{T}_{v,g}} := \text{diag}\{\sigma^2_1, \sigma^2_2, \sigma^2_3, \sigma^2_4, \sigma^2_5, \sigma^2_6\}$, where $\sigma_1 \dots \sigma_6$ are tunable parameters corresponding to the six generators of SE(3).
Together these factors form an 8-dimensional Gaussian distribution with covariance $\Vec{R}_i := \text{diag}\{ \Vec{R}_{\Vec{y}^i}, \Vec{R}_{\Vec{T}_{v,g}} \}$, which we propagate via the combined Jacobian
\begin{equation*} \label{eq:jacobian}
\Vec{G}_i := \bbm    \dfrac{\partial\Vec{z}^i_c}{\partial\Vec{y}^i}  & \dfrac{\partial\Vec{z}^i_c}{\partial \Vec{T}_{v,g}}
       \ebm
\end{equation*}
to approximate $\Vec{z}^i_c$ as a Gaussian in 3D space with covariance $\Vec{Q}_i = \Vec{G}_i\Vec{R}_i\Vec{G}_i^T$.

Using the Cartesian coordinates of $\Vec{z}^i_c$ and $\Vec{y}^i$ to compute the Jacobian, we have
\begin{align}
\frac{\partial\Vec{z}^i_c}{\partial\Vec{y}^i} &= \frac{z^i_c}{k_1}
\bbm
\left(k_1+k_3 x^i_c\right) / f_u & k_4 x^i_c / f_v \\[0.5ex]
k_3 y^i_c / f_u & \left(k_1 + k_4 y^i_c\right) / f_v \\[0.5ex]
k_3 z^i_c / f_u & k_4 z^i_c / f_v
\ebm
\end{align}
and
\begin{align}
\frac{\partial\Vec{z}^i_c}{\partial{\Vec{T}_{v,g}}} &= \frac{\partial\Vec{z}^i_c}{\partial{\Vec{T}_{c,g}}} \frac{\partial \Vec{T}_{c,g}}{\partial{\Vec{T}_{v,g}}} = \bbm \Vec{1} & (-\Vec{z}^i_c)^\times \ebm \text{Ad}(\Vec{T}_{c,v}) \text{ .}
\end{align}
In the above, we adopt the notation of \cite{Barfoot2014}: $\Vec{1}$ denotes the $(3\times3)$ identity matrix, $\text{Ad}(\cdot)$ the adjoint in SE(3), and $(\cdot)^\times$ the skew-symmetric cross-product matrix.

Figure \ref{fig:ellipses} shows $1 \sigma$ uncertainty ellipses for a number of evenly spaced synthetic image features resulting from a camera configuration similar to that used in the experiments described in Section \ref{sec:experiments}.

\section{System Overview} \label{sec:system}
\begin{figure}[b]
    \centering
    \includegraphics[width=\textwidth]{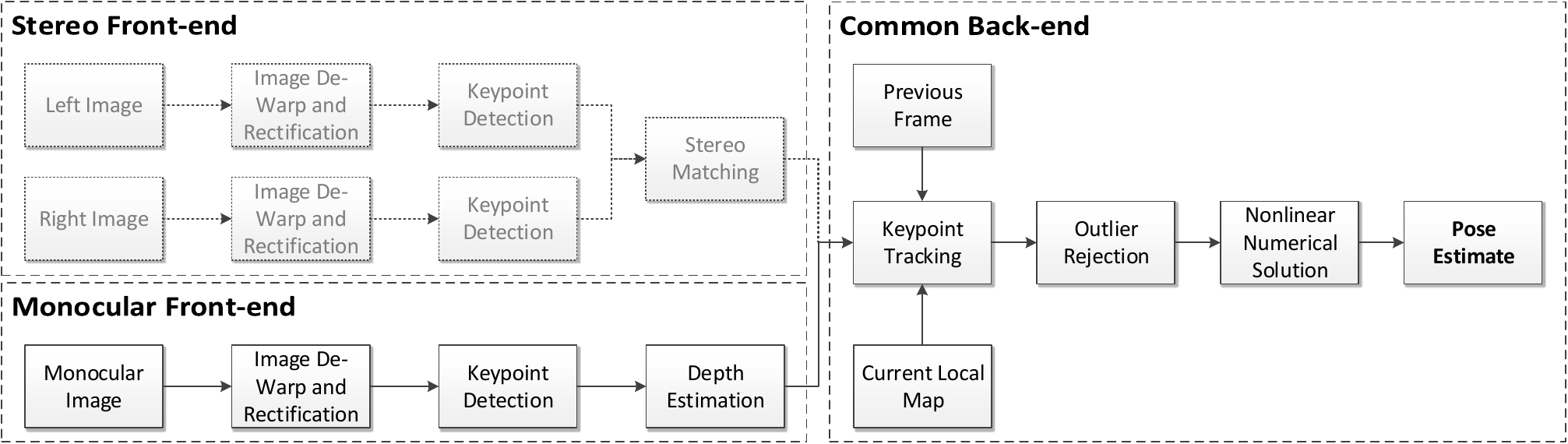}
    \caption{The major processing blocks of the stereo and monocular localization pipelines. The monocular pipeline shares most of the same processing blocks as its stereo counterpart, differing mainly in the front-end image processing used to generate 3D keypoints. The ``Current Local Map'' block is only used for keypoint tracking during the repeat pass.}
    \label{fig:pipeline}
\end{figure}

This section provides an overview of the VT\&R system as it pertains to the methods of the previous section.
In particular, we discuss the generic localization pipeline used for both online mapping in the teach pass and local map construction in the repeat pass.
Figure \ref{fig:pipeline} shows the stereo and monocular versions of the pipeline, which differ mainly in the front-end image processing used to generate 3D keypoints.

\subsection{Keypoint Generation}
Raw images entering the pipeline first pass through a pre-processing step that uses a calibrated camera model to make them appear as though they were produced by an ideal pinhole camera.
A GPU implementation of the SURF detector \cite{Bay2008} then identifies keypoints in the de-warped and rectified images. 
The pipeline estimates the 3D coordinates of each keypoint in the camera frame using a matching procedure in the stereo case or the technique of Section \ref{sec:depth} in the monocular case.
The subsequent behavior of the pipeline differs slightly between the teach pass and the repeat pass.

\subsection{Teach Pass}
In the teach pass, the system constructs a pose graph whose vertices store lists of 3D keypoints with associated uncertainty and SURF descriptors, and whose edges store lists of matched keypoints and 6DOF pose change estimates.
The system first tracks 3D keypoints in the current image against those in the most recent keyframe to generate a list of keypoint matches.
These matches form the input to a 3-point RANSAC algorithm \cite{Fischler1981} that generates hypotheses for the 6DOF interframe pose change and rejects outlying feature tracks.
In the context of monocular VT\&R, this procedure typically rejects features far from the local ground surface (e.g., walls) since their motion is not adequately captured by the uncertainty model described in Section \ref{sec:uncertainty}.
The resulting pose change estimate serves as the initial guess in an iterative Gauss-Newton   that refines the estimate based on inlying tracks.

\subsection{Repeat Pass}
The repeat pass begins with a manual initialization at some vertex in the pose graph, and the specification of a destination vertex.
The system then reconstructs the vehicle's path from the appropriate chain of relative transformations.

At every timestep, the system identifies the nearest keyframe in the path and performs a local bundle adjustment over a user-specified number of topologically adjacent keyframes, generating a local metric map in the reference frame of the nearest keyframe.
The system then forms an augmented keyframe from the adjusted map keypoints against which freshly detected features may be matched.
As in the teach pass, the system performs frame-to-frame VO to obtain an initial 6DOF pose estimate at each time step, which it uses as an initial guess to localize against the current local map and refine its pose estimate.

If the system fails to localize against the map, it may rely purely on VO until either a successful localization occurs or the vehicle exceeds some preset distance threshold since the last successful localization.
In the latter case, the system will halt the traverse and enter a search mode until it relocalizes or the operator intervenes.

\section{Experiments} \label{sec:experiments}
We conducted two sets of experiments at the University of Toronto Institute for Aerospace Studies (UTIAS), the first outdoors on relatively flat terrain, and the second on the highly non-planar terrain of the UTIAS MarsDome indoor rover testing environment.
We compare the performance of our monocular VT\&R system to that of the established stereo system \cite{Furgale2010} over 4.3 km of autonomous navigation.
Table \ref{table:results} reports path lengths, repeat speeds, start times, and autonomy rates for each experiment.
We repeated each route using the monocular pipeline first, and conducted each experiment between roughly 10:00 and 14:00 when the sun was highest in the sky to minimize the effects of lighting changes and shadows.

\begin{table}[b]
    \centering
    \caption{Summary of experimental results}
    \begin{threeparttable}
        \begin{tabular}{cccccccccccc}
            &&&&& \multicolumn{3}{c}{Local start time (UTC-4)} && \multicolumn{3}{c}{Autonomy rate} \B \\ \cline{6-8} \cline{10-12}
            Trial & Route & Path length & Repeat speed  && Teach & Mono & Stereo &&& Mono & Stereo \T\B \\ \hline
            1 & Outdoor & 1370 m & 0.6 m/s  && 09:56:46 & 10:35:10 & 12:08:30 &&& 99.71\%\tnote{\dag} \quad & 100.00\% \T \\
            2 & Outdoor & 1360 m & 0.6 m/s  && 11:45:40 & 12:22:26 & 13:43:49 &&& 99.88\% & 100.00\% \\
            3 & Outdoor & 1361 m & 0.6 m/s  && 13:26:41 & 14:00:12 & 15:20:12 &&& 99.74\% & 100.00\% \\
            4 & Indoor & 126 m & 0.3 m/s  && 13:32:23 & 13:40:53 & 14:02:46 &&& 96.28\% & 100.00\%  \\
            5 & Indoor & 140 m & 0.3 m/s  && 12:18:57 & 12:32:20 & 12:59:11 &&& 91.60\% & 100.00\% \B \\ \hline
            &&&&& \textbf{Mono} & \textbf{Stereo} &&& \T \B \\
            \multicolumn{5}{l}{\textbf{Total distance driven}} & 4298 m\tnote{\dag} & 4357 m &&&&& \\
            \multicolumn{5}{l}{\textbf{Total distance autonomously traversed}} & 99.41\% & 100.00\% &&&&& \B \\ \hline
        \end{tabular}
        \begin{tablenotes}
            \item [\dag] During the monocular repeat pass of Trial 1, a parked vehicle on the path forced manual driving for 59 m before successful relocalization. We exclude this segment in our analysis and report the monocular autonomy rate for Trial 1 based on a reduced path length of 1311 m.
        \end{tablenotes}
    \end{threeparttable}
    \label{table:results}
\end{table}

\subsection{Hardware}
We used a four-wheeled skid-steered Clearpath Husky A200 rover equipped with a PointGrey Bumblebee XB3 stereo camera, which outputs $512\times384$ pixel greyscale images at 15 frames per second.
The camera is mounted 1.0 m above the ground and is angled downwards at $47^\circ$ to the horizontal.
These values were measured by hand since our system functions well even without an especially accurate estimate of $\Vec{T}_{c,v}$.
Small errors in $\Vec{T}_{c,v}$ are simply absorbed by the uncertainty in $\Vec{T}_{v,g}$.

\begin{figure}
	\sidecaption[t]
	\includegraphics[width=0.45\textwidth]{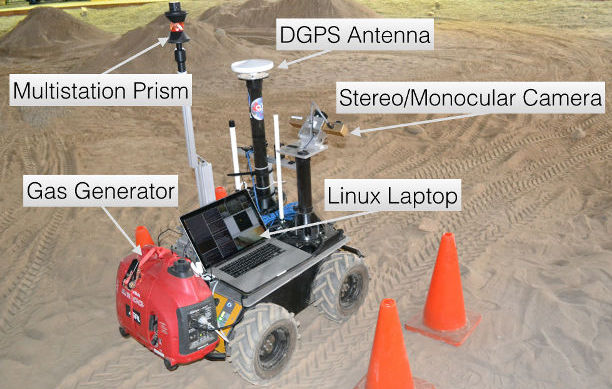}
	\caption{Clearpath Husky A200 rover equipped with PointGrey Bumblebee XB3 stereo camera, DGPS receiver, Leica Nova MS50 MultiStation prism, 1 kW gas generator, and Linux laptop running ROS \cite{Quigley2009} .}
\end{figure}

During the teach pass, we recorded stereo images and used them to teach identical paths using both the monocular and stereo pipelines.
For the monocular pipeline, we used imagery from the left camera of the stereo pair only.
The system detects 600 SURF keypoints in each incoming image and creates new keyframes every 25 cm in translation or $2.5^\circ$ in rotation.
For the monocular pipeline, we assigned standard deviations of 10 cm to the translational components of $\Vec{T}_{v,g}$ and $10^\circ$ to its rotational components as these values generally worked well in practice.

\subsection{Outdoor Experiments}
\begin{figure}[b]
	\centering
	\includegraphics[width=0.95\textwidth]{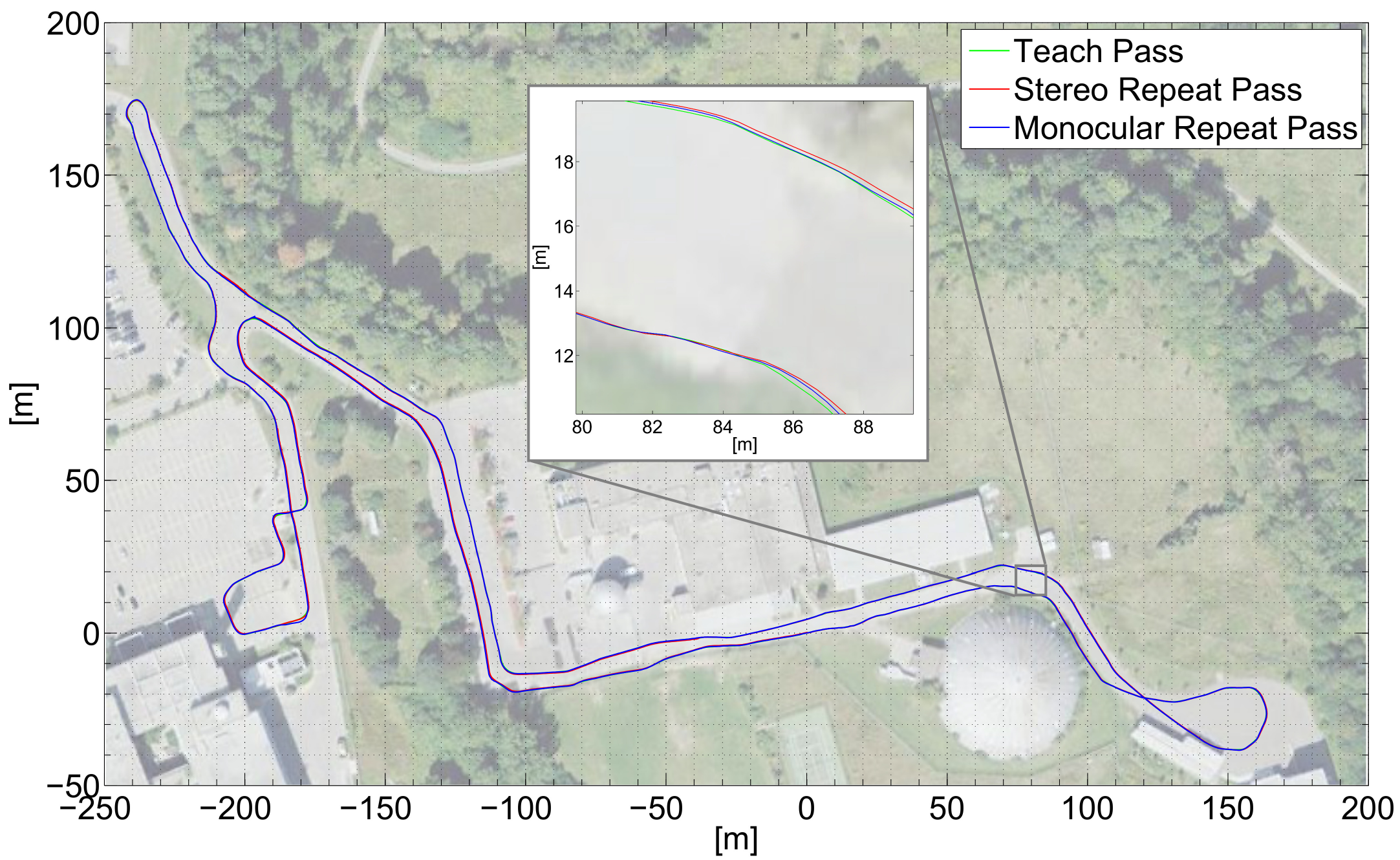}
	\caption{Comparison of RTK-corrected GPS tracks of the teach pass, stereo repeat pass, and monocular repeat pass of a 1.4 km outdoor route (Trial 3 in Table \ref{table:results}). The zoomed-in section highlights the centimetre-level accuracy of both pipelines. (Map data: Google, DigitalGlobe.)}
	\label{fig:outdoor_gps}
\end{figure}

To evaluate the performance of the monocular VT\&R system over long distances, we taught three 1.4 km paths through the parking lots and driveways of UTIAS.
While these paths consisted mostly of flat pavement, they included many non-planar features such as speed bumps, side slopes, deep puddles, and rough shoulders, as well as other terrain types including gravel, sand, and grass.

We equipped the rover with an Ashtech DG14 Differential GPS unit used in tandem with a second stationary DG14 unit to obtain centimetre-accuracy RTK-corrected GPS data during the outdoor experiments.
We used these data purely for ground-truthing purposes; they had no bearing on the behaviour of either pipeline.
Figure \ref{fig:outdoor_gps} shows GPS tracks of the teach and repeat passes of one outdoor route.

\begin{figure}
    \centering
    \subfigure[Monocular repeat pass] {
        \includegraphics[width=0.43\textwidth]{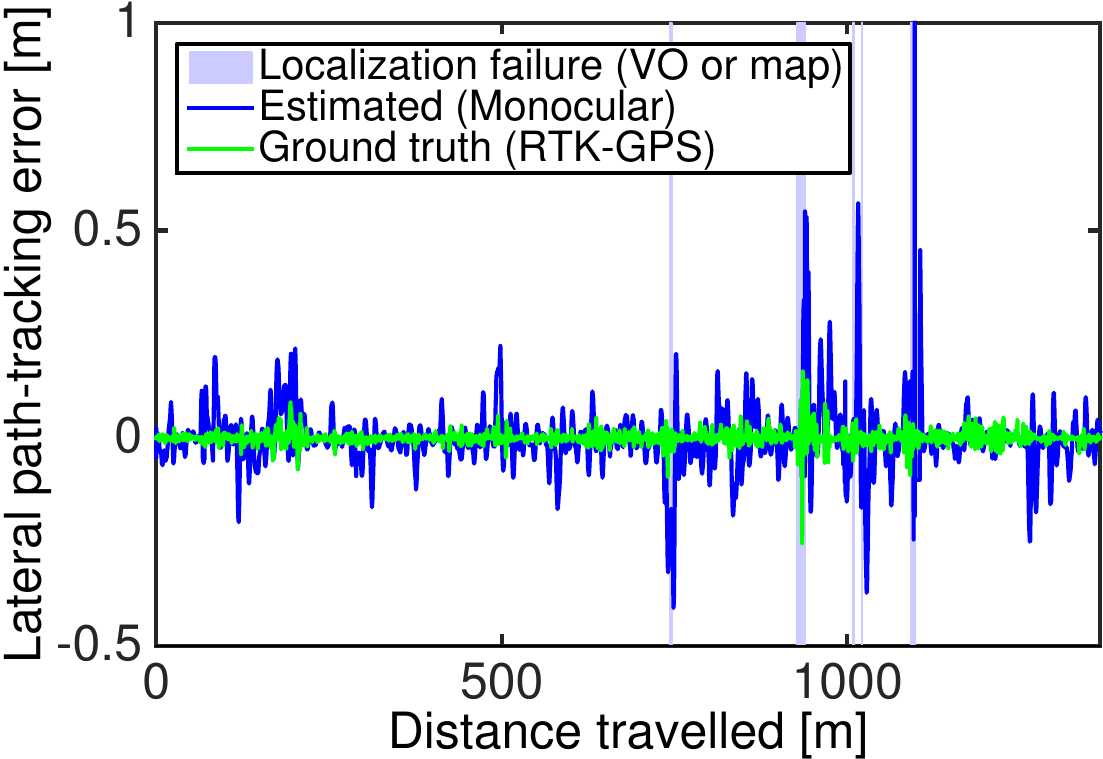}
        \label{fig:outdoortracking_mono} }
    ~
    \subfigure[Stereo repeat pass] {
        \includegraphics[width=0.43\textwidth]{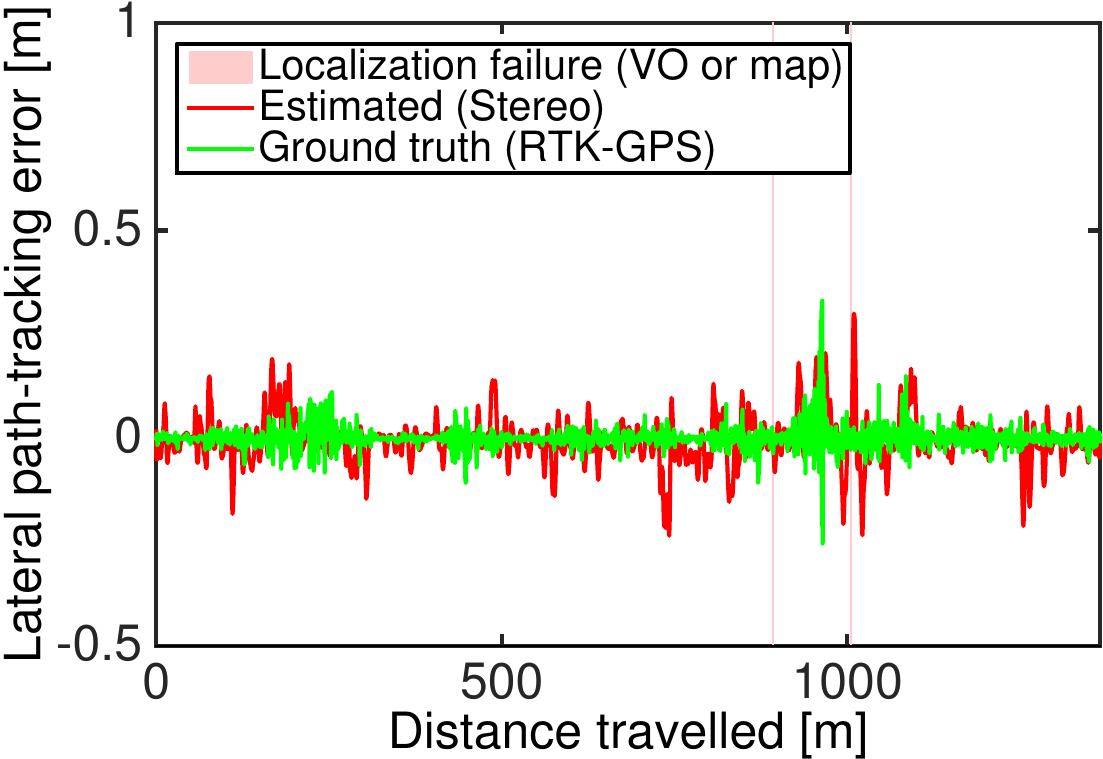}
        \label{fig:outdoortracking_stereo} }
    \caption{Estimated and measured lateral path-tracking error during the monocular and stereo repeat passes of the 1.4 km outdoor route shown in Figure \ref{fig:outdoor_gps} (Trial 3 in Table \ref{table:results}). GPS tracking shows that both monocular and stereo VT\&R achieve centimetre-level accuracy, although estimated lateral path-tracking error tends to diverge from the true value in cases of localization failure.}
    \label{fig:outdoortracking}
\end{figure}

\begin{figure}
    \centering
    \subfigure[VO feature matches] {
        \includegraphics[width=0.43\textwidth]{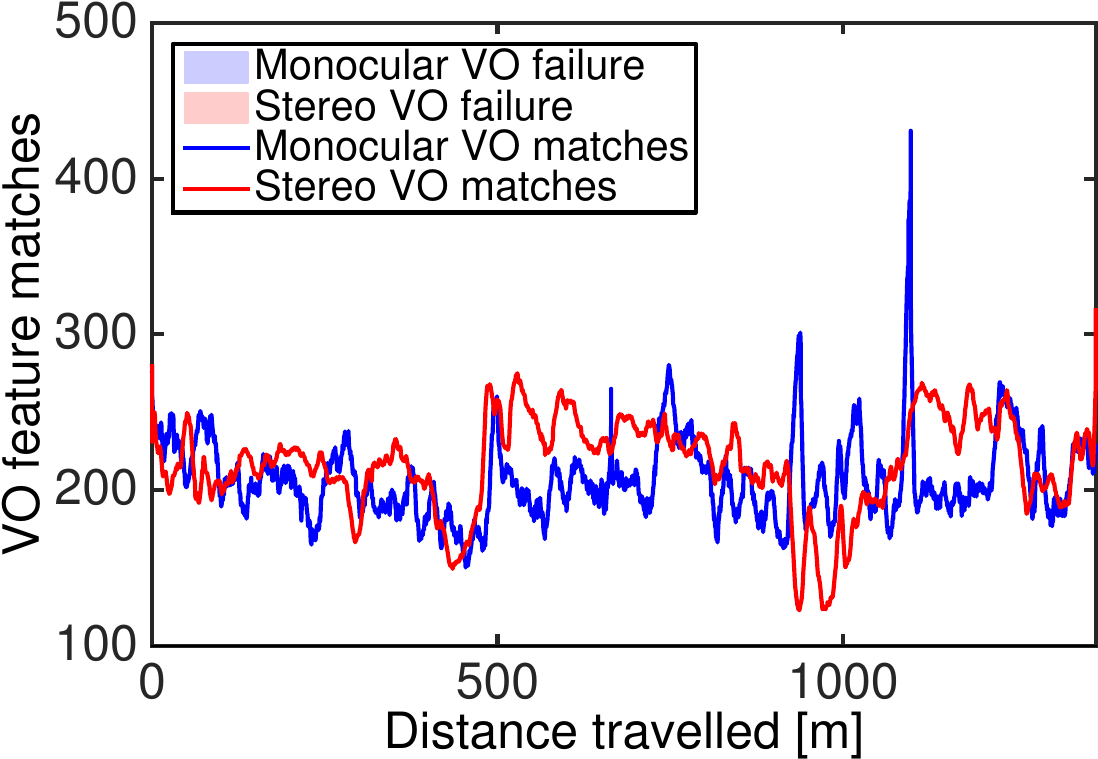}
        \label{fig:outdoor_vo_matches} }
    ~
    \subfigure[Map feature matches] {
        \includegraphics[width=0.43\textwidth]{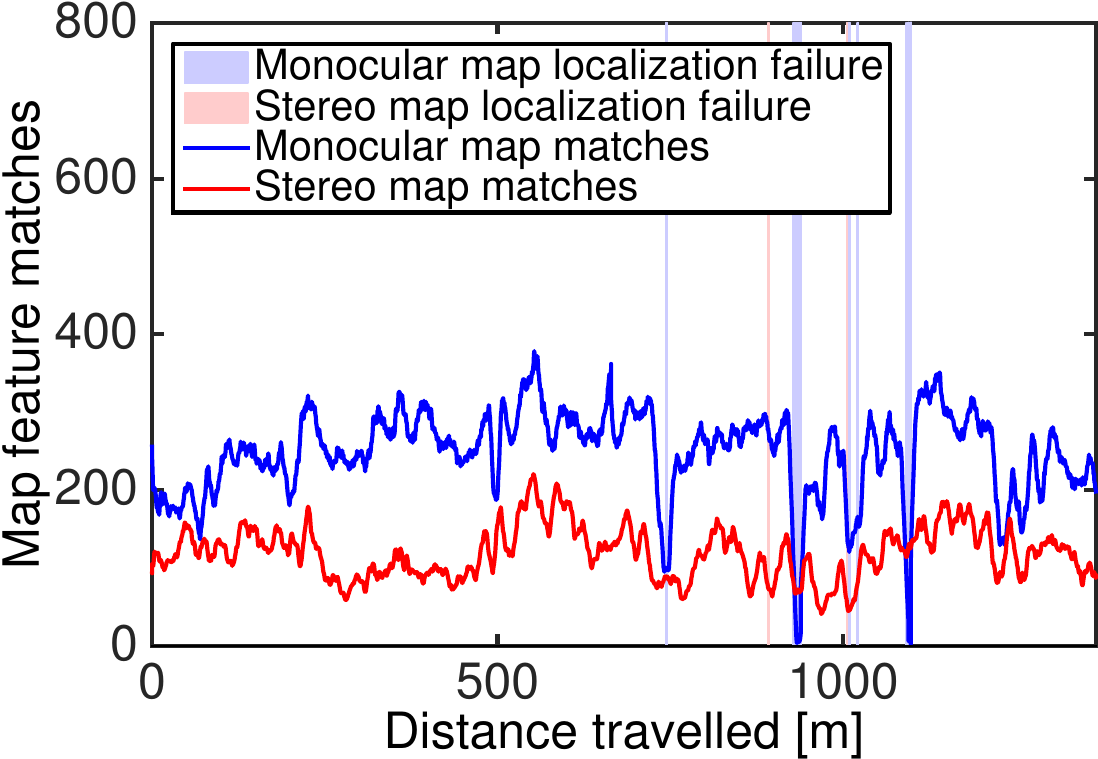}
        \label{fig:outdoor_map_matches} }
    
    \caption{Keypoint matches during the monocular and stereo repeat passes of the 1.4 km outdoor route shown in Figure \ref{fig:outdoor_gps} (Trial 3 in Table \ref{table:results}), with localization failures highlighted. A localization failure is defined as less than 10 feature matches. There were no VO failures during either repeat pass. For clarity, we have applied a 20-point sliding-window mean filter to the raw data.}
    \label{fig:outdoormatches}
\end{figure}

Figure \ref{fig:outdoortracking} shows estimated and measured lateral path-tracking errors during the monocular and stereo repeat passes.
Both pipelines achieved centimetre-level accuracy in their respective repeat passes and produced similar estimates of lateral path-tracking error.
In cases of map localization failure (i.e., when the system relied on pure VO), the monocular pipeline's estimated lateral path-tracking error diverged from ground truth more quickly than that of the stereo pipeline since keypoint position uncertainties are poorly constrained by only two measurements.
Note, however, that the vehicle remained within about 20 cm of the taught path at all times.

Figure \ref{fig:outdoormatches} compares the number of successful feature matches for frame-to-frame VO and map-based localization for both pipelines.
Both pipelines track similar numbers of features from frame to frame, but the monocular pipeline generally tracks twice as many map features as its stereo counterpart.
This result is most likely due to bad data association during local map construction in the monocular pipeline, which stems from the comparatively large positional uncertainties of distant keypoints.

Bad data association is especially problematic in regions of highly self-similar terrain (e.g., Figure \ref{fig:textureless}) since large positional uncertainties exacerbate ambiguity in feature matches.
With fewer correctly associated measurements, the bundle adjustment procedure will not maximally constrain the positions of map keypoints, which we would expect to increase the risk of localization failures.
Indeed, Figure \ref{fig:outdoor_map_matches} shows that the monocular pipeline suffered more serious map localization failures than the stereo pipeline, although these forced manual intervention only once.

\subsection{Indoor Experiments}

\begin{figure}[b]
	\centering
	\includegraphics[width=0.95\textwidth]{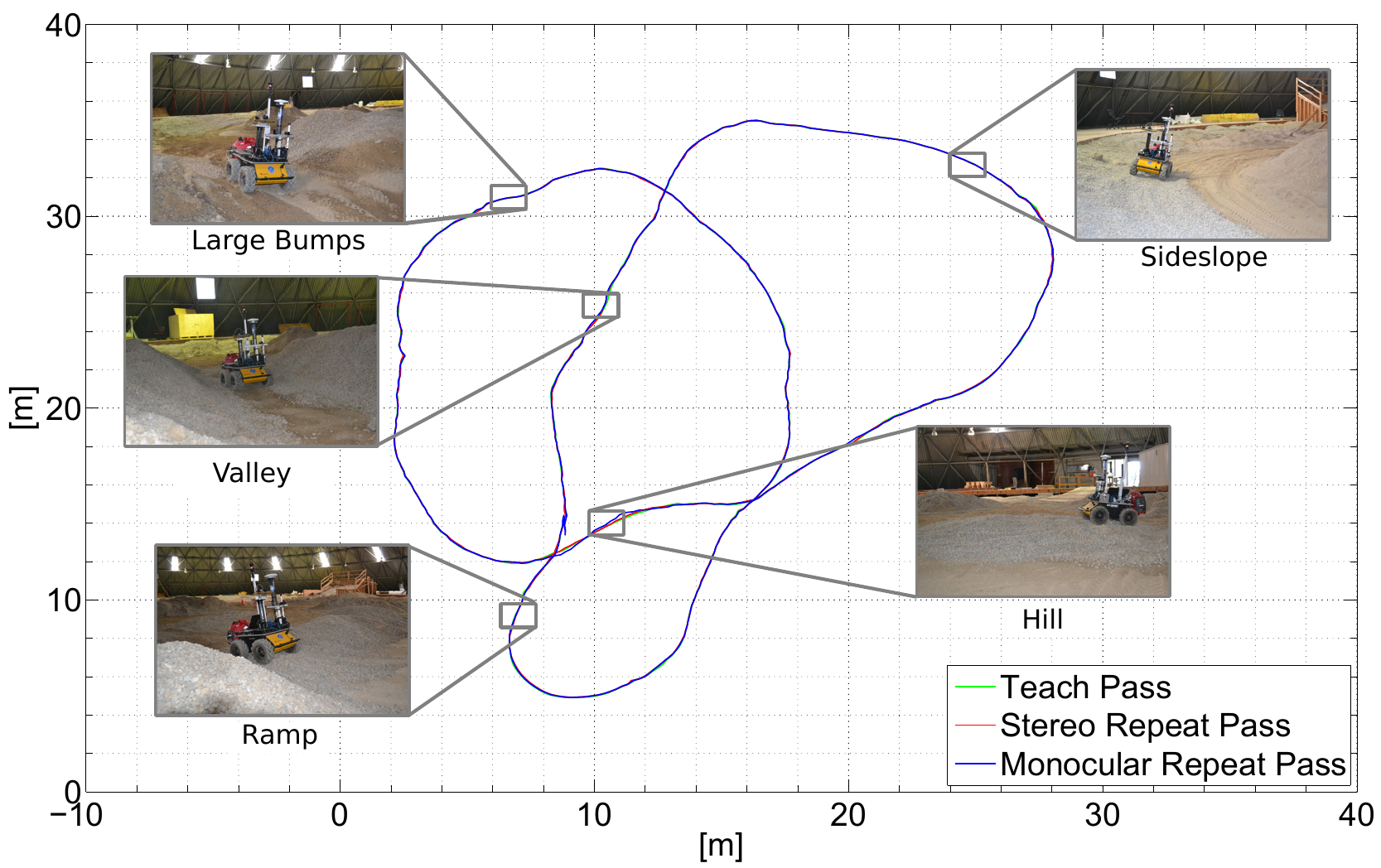}
	\caption{Comparison of MultiStation tracks of the teach pass, stereo repeat pass, and monocular repeat pass of a 140 m indoor route (Trial 5 in Table \ref{table:results}), with some interesting segments highlighted.}
	\label{fig:indoorroute}
\end{figure}

The second set of experiments took place in the more challenging terrain of the UTIAS MarsDome.
These routes included a number of highly non-planar features such as hills, large bumps, valleys, and slopes of a similar scale to the vehicle.

Since the MarsDome is an enclosed facility, GPS tracking was not available, and we instead made use of a Leica Nova MS50 MultiStation to track the position of the rover with millimetre-level accuracy.
Similarly to the outdoor experiments, we used these data for ground-truthing purposes only.
Figure \ref{fig:indoorroute} shows MultiStation data of the teach and repeat passes of a 140 m route through the MarsDome, along with images of some of the more challenging terrain features on the route. 

\begin{figure}
    \centering
    \subfigure[Monocular repeat pass] {
        \includegraphics[width=0.43\textwidth]{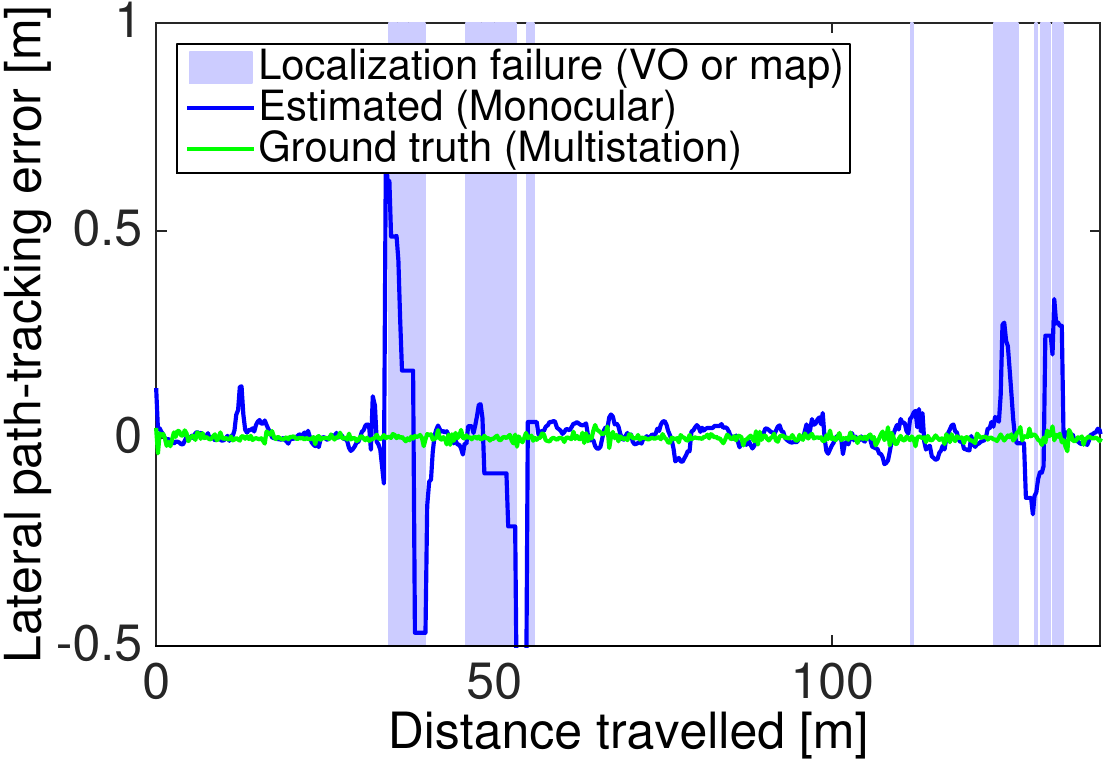} }
    ~
    \subfigure[Stereo repeat pass] {
        \includegraphics[width=0.43\textwidth]{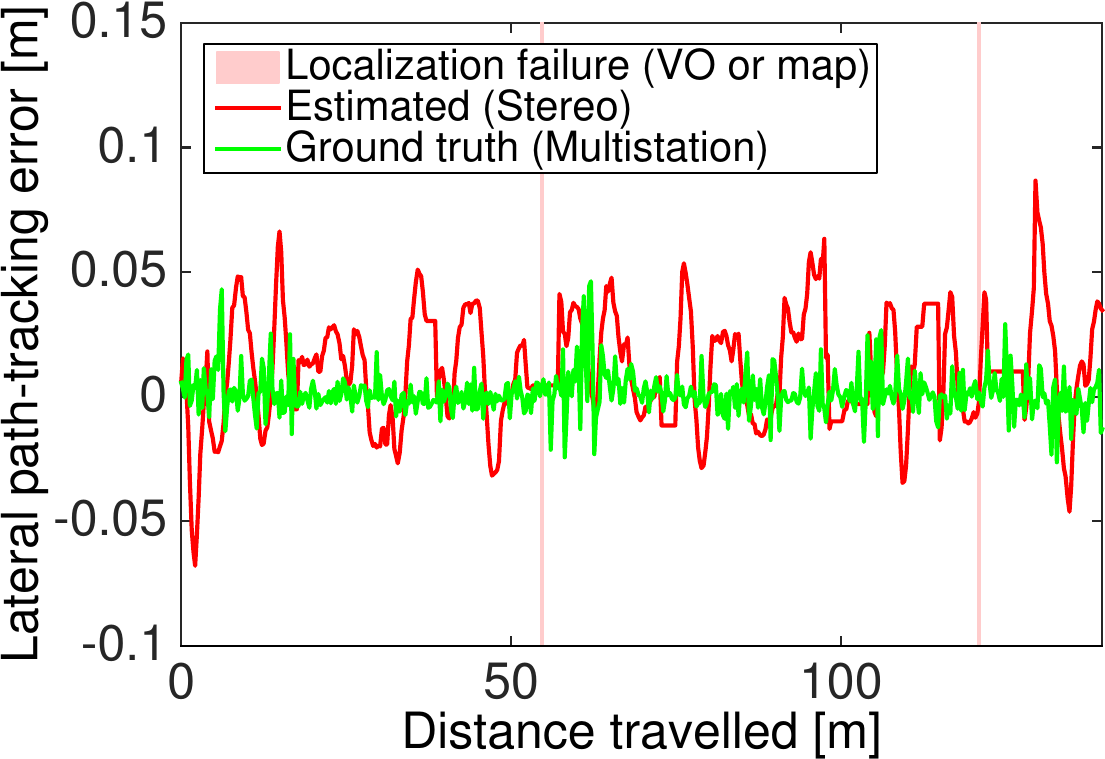} }
    
    \caption{Estimated and measured lateral path-tracking error during the monocular and stereo repeat passes of the 140 m indoor route shown in Figure \ref{fig:indoorroute} (Trial 5 in Table \ref{table:results}). MultiStation tracking shows that both monocular and stereo VT\&R achieve centimetre-level accuracy in highly non-planar terrain, although estimated lateral path-tracking error tends to diverge from the true value in cases of localization failure. Note the difference in scale between the two plots.}
    \label{fig:indoortracking}
\end{figure}

\begin{figure}
    \centering
    \subfigure[VO features matches] {
        \includegraphics[width=0.43\textwidth]{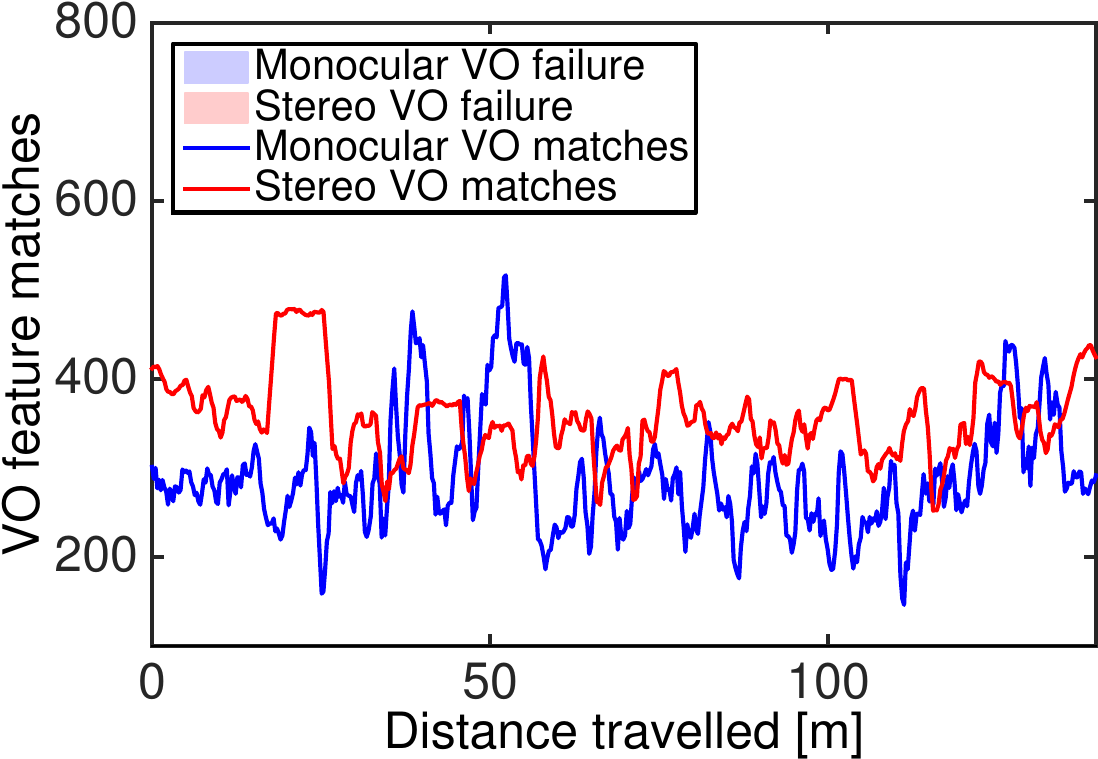} }
    ~
    \subfigure[Map feature matches] {
        \includegraphics[width=0.43\textwidth]{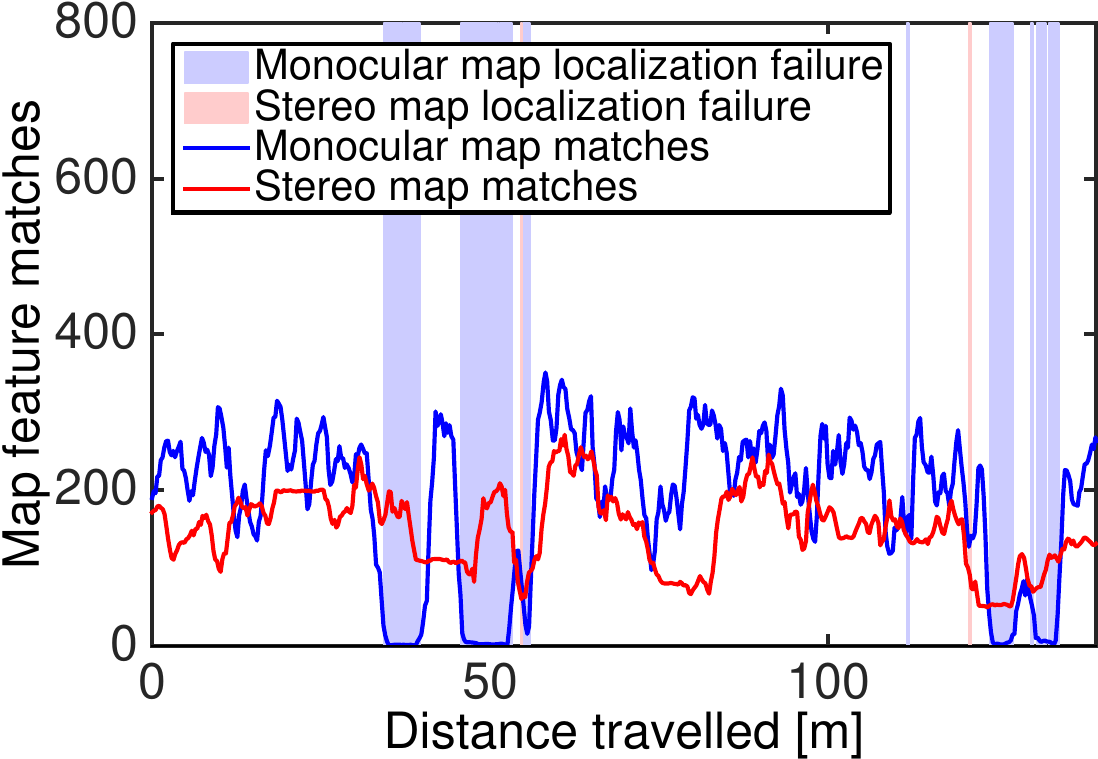} }
    
    \caption{Keypoint matches during the monocular and stereo repeat passes of the 140 m indoor route shown in Figure \ref{fig:indoorroute} (Trial 5 in Table \ref{table:results}), with localization failures highlighted. A localization failure is defined as less than 10 feature matches. There were no VO failures during either repeat pass. For clarity, we have applied a 5-point sliding-window mean filter to the raw data.}
    \label{fig:indoormatches}
\end{figure}

\begin{figure}
    \centering
	\subfigure[Self-similar terrain] {
    	\includegraphics[width=0.4\textwidth]{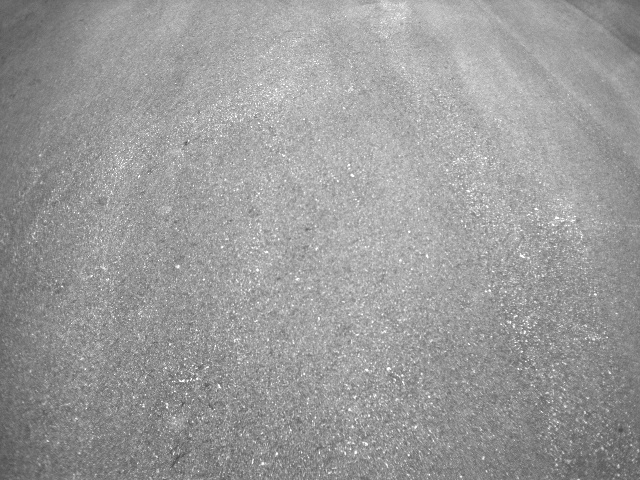}
    	\label{fig:textureless}
	}
    \subfigure[Motion blur] {
    	\includegraphics[width=0.4\textwidth]{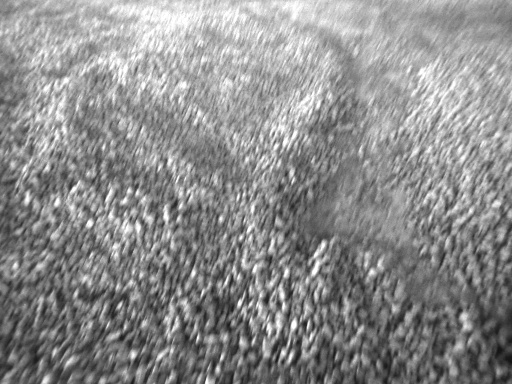}
    	\label{fig:motionblur}
	}
    \caption{The most common causes of localization failure were highly self-similar terrain and motion blur. Neither stereo nor monocular VT\&R is immune to these conditions, but their effects were exacerbated by high spatial uncertainty in the monocular case.}
\end{figure}

Figure \ref{fig:indoortracking} shows estimated and measured lateral path-tracking errors for the monocular and stereo repeat passes.
As in the outdoor case, both pipelines achieved centimetre-level accuracy, even in difficult terrain.
Again, note that although the monocular pipeline's estimated lateral path-tracking error diverged significantly from ground-truth during localization failures, the MultiStation tracks show that the vehicle remained within a few centimetres of the path throughout the traverse.

Figure \ref{fig:indoormatches} shows VO and map feature matches for both repeat passes.
The monocular pipeline suffered map localization failures more often than the stereo pipeline, the worst failure occurring in the valley and hill regions (see Figure \ref{fig:indoorroute}) where the lighting was especially poor.
This led to increased motion blur (see Figure \ref{fig:motionblur}) and poor feature matching due to greater uncertainty in keypoint positions.
Both failures necessitated manual intervention over a few metres, however, the system successfully relocalized once the lighting improved.

\section{Lessons Learned and Future Work} \label{sec:lessons}
Experiments with our systems led to several useful lessons and possible extensions:
\begin{enumerate}
	\item With sufficient spatial uncertainty, the flat-ground assumption seems to be usable even in rough driving conditions, provided the scene is well-lit and reasonably textured. Steep hills were problematic for monocular VT\&R since the camera would observe features mainly on the horizon or on walls during the ascent.
	\item The performance our systems depends on a search (often manual) through a high-dimensional space of tuning parameters, and it is difficult to be certain that an optimal configuration has been found. Iterative learning algorithms such as \cite{Ostafew2013} may present a solution by learning optimal parameters from experience.
	\item Data association quality is not a monotonic function of observation uncertainty. Too little uncertainty and good feature matches get rejected; too much and all matches are equally good (or bad). Both cases result in tracking failure. This reinforces the need for an accurate model of a system's noise properties.
	\item Experimenting with camera orientation could improve the accuracy of monocular VT\&R, particularly on hills. For example, orienting the camera perpendicular to the direction of travel has been shown to improve the accuracy of stereo visual odometry \cite{Peretroukhin2014}.
	\item By using stereo vision in the teach pass and monocular vision in the repeat pass, we could forgo the flat-ground assumption for mapping, which should result in fewer localization failures in the repeat pass.
\end{enumerate}
%

\section{Conclusions} \label{sec:conclusions}
This paper has described a Visual Teach and Repeat (VT\&R) system capable of autonomously repeating kilometre-scale routes in rough terrain using only monocular vision.
By constraining features of interest to lie on a manifold of uncertain local ground planes, we relax the requirement for true 3D sensing that had prevented the deployment of Furgale and Barfoot's VT\&R system \cite{Furgale2010} on a wide range of vehicles equipped with monocular cameras.
Extensive field tests have demonstrated that this system is capable of achieving centimetre-level accuracy on par with its stereo counterpart, but that there is an associated trade-off in robustness.
Nevertheless, we believe that the benefit of deploying VT\&R on existing vehicles without requiring the installation of additional sensors far outweighs the associated reduction in robustness.

\begin{acknowledgement}
The authors would like to thank Matthew Giamou and Valentin Peretroukhin of the Space and Terrestrial Autonomous Robotic Systems (STARS) lab for their assistance with field testing, the Autonomous Space Robotics Lab (ASRL) for their guidance in interacting with the VT\&R code base, Leica Geosystems for providing the MultiStation, and Clearpath Robotics for providing the Husky rover. This work was supported by the Natural Sciences and Engineering Research Council (NSERC) through the NSERC Canadian Field Robotics Network (NCFRN).
\end{acknowledgement}

\newpage
\def\url#1{}
\def\doi#1{}
\bibliographystyle{spmpsci}

\begin{thebibliography}{10}
\providecommand{\url}[1]{{#1}}
\providecommand{\urlprefix}{}
\expandafter\ifx\csname urlstyle\endcsname\relax
  \providecommand{\doi}[1]{DOI~\discretionary{}{}{}#1}\else
  \providecommand{\doi}{DOI~\discretionary{}{}{}\begingroup
  \urlstyle{rm}\Url}\fi

\bibitem{Barfoot2014}
Barfoot, T., Furgale, P.: Associating uncertainty with three-dimensional poses
  for use in estimation problems.
\newblock IEEE Trans. Robot. (T-RO) \textbf{30}(3), 679--693 (2014)

\bibitem{Bay2008}
Bay, H., Ess, A., Tuytelaars, T., Gool, L.V.: {Speeded-up robust features
  (SURF)}.
\newblock Comput. Vision and Image Understanding (CVIU) \textbf{110}, 346--359
  (2008).
\newblock
  \urlprefix\url{http://www.sciencedirect.com/science/article/pii/S1077314207001555}

\bibitem{Choi2011}
Choi, S., Joung, J., Yu, W., Cho, J.: {Monocular visual odometry under planar
  motion constraint}.
\newblock Proc. Int. Conf. Control, Autom. and Sys. (ICCAS) pp. 1480--1485
  (2011).
\newblock
  \urlprefix\url{http://ieeexplore.ieee.org/xpls/abs\_all.jsp?arnumber=6106228}

\bibitem{Davison2007}
Davison, A.J., Reid, I.D., Molton, N.D., Stasse, O.: {MonoSLAM: Real-time
  single camera SLAM}.
\newblock IEEE Trans. Pattern Anal. Mach. Intell. (TPAMI) \textbf{29}(6),
  1052--1067 (2007).
\newblock
  \urlprefix\url{http://ieeexplore.ieee.org/xpls/abs\_all.jsp?arnumber=4160954}

\bibitem{Eade2006}
Eade, E., Drummond, T.: {Scalable monocular SLAM}.
\newblock In: Proc. IEEE Conf. Comput. Vision and Pattern Recognition (CVPR)
  (2006).
\newblock
  \urlprefix\url{http://ieeexplore.ieee.org/xpls/abs\_all.jsp?arnumber=1640794}

\bibitem{Farraj2013}
Farraj, F., Asmar, D.: {Non-iterative planar visual odometry using a monocular
  camera}.
\newblock In: Proc. Int. Conf. Advanced Robot. (ICAR), pp. 1--6 (2013).
\newblock
  \urlprefix\url{http://ieeexplore.ieee.org/xpls/abs\_all.jsp?arnumber=6766475}

\bibitem{Fischler1981}
Fischler, M.A., Bolles, R.C.: {Random sample consensus: a paradigm for model
  fitting with applications to image analysis and automated cartography}.
\newblock Commun. ACM \textbf{24}(6) (1981).
\newblock \urlprefix\url{http://dl.acm.org/citation.cfm?id=358692}

\bibitem{Furgale2010}
Furgale, P., Barfoot, T.D.: {Visual teach and repeat for long-range rover
  autonomy}.
\newblock J. Field Robot. (JFR) \textbf{27}(5), 534--560 (2010).
\newblock
  \urlprefix\url{http://onlinelibrary.wiley.com/doi/10.1002/rob.20342/full}

\bibitem{Goedeme2007}
Goedem\'{e}, T., Nuttin, M., Tuytelaars, T., Gool, L.V.: {Omnidirectional
  vision based topological navigation}.
\newblock Int. J. Comput. Vision (IJCV) \textbf{74}(3), 219--236 (2007).
\newblock
  \urlprefix\url{http://link.springer.com/article/10.1007/s11263-006-0025-9}

\bibitem{Holmes2013}
Holmes, S.A., Murray, D.W.: {Monocular SLAM with Conditionally Independent
  Split Mapping}.
\newblock IEEE Trans. Pattern Anal. Mach. Intell. (TPAMI) \textbf{35}(6),
  1451--1463 (2013).
\newblock
  \urlprefix\url{http://ieeexplore.ieee.org/xpls/abs\_all.jsp?arnumber=6341750}

\bibitem{Kidono2002}
Kidono, K., Miura, J., Shirai, Y.: {Autonomous visual navigation of a mobile
  robot using a human-guided experience}.
\newblock Robot. and Autonomous Syst. (RAS) \textbf{40}(2-3), 121--130 (2002).
\newblock
  \urlprefix\url{http://linkinghub.elsevier.com/retrieve/pii/S0921889002002373}

\bibitem{Klein2007}
Klein, G., Murray, D.: {Parallel tracking and mapping for small AR workspaces}.
\newblock In: Proc. IEEE/ACM Int. Symp. Mixed and Augmented Reality (ISMAR)
  (2007).
\newblock
  \urlprefix\url{http://ieeexplore.ieee.org/xpls/abs\_all.jsp?arnumber=4538852}

\bibitem{Lovegrove2011}
Lovegrove, S., Davison, A.J., Ibanez-Guzman, J.: {Accurate visual odometry from
  a rear parking camera}.
\newblock In: Proc. Intelligent Vehicles Symp. (IV) (2011).
\newblock
  \urlprefix\url{http://ieeexplore.ieee.org/xpls/abs\_all.jsp?arnumber=5940546}

\bibitem{Marshall2008}
Marshall, J., Barfoot, T.D., Larsson, J.: {Autonomous underground tramming for
  center-articulated vehicles}.
\newblock J. Field Robot. (JFR) \textbf{25}, 400--421 (2008).
\newblock
  \urlprefix\url{http://onlinelibrary.wiley.com/doi/10.1002/rob.20242/abstract}

\bibitem{Matsumoto1996}
Matsumoto, Y., Inaba, M., Inoue, H.: {Visual navigation using view-sequenced
  route representation}.
\newblock In: Proc. IEEE Int. Conf. Robot. and Autom. (ICRA), pp. 83--88
  (1996).
\newblock
  \urlprefix\url{http://ieeexplore.ieee.org/xpls/abs\_all.jsp?arnumber=503577}

\bibitem{McManus2013}
McManus, C., Furgale, P., Stenning, B., Barfoot, T.D.: {Lighting-invariant
  Visual Teach and Repeat Using Appearance-based Lidar}.
\newblock J. Field Robot. (JFR) \textbf{30}(2), 254--287 (2013).
\newblock
  \urlprefix\url{http://onlinelibrary.wiley.com/doi/10.1002/rob.21444/full}

\bibitem{Ostafew2013}
Ostafew, C., Schoellig, A., Barfoot, T.: Iterative learning control to improve
  mobile robot path tracking in challenging outdoor environments.
\newblock In: Proc. IEEE/RSJ Int. Conf. Intelligent Robot. and Syst. (IROS),
  pp. 176--181 (2013)

\bibitem{Peretroukhin2014}
Peretroukhin, V., Kelly, J., Barfoot, T.: Optimizing camera perspective for
  stereo visual odometry.
\newblock In: Proc. Conf. Comput. and Robot Vision (CRV), pp. 1--7 (2014)

\bibitem{Quigley2009}
Quigley, M., Conley, K., Gerkey, B.P., Faust, J., Foote, T., Leibs, J.,
  Wheeler, R., Ng, A.Y.: {ROS: an open-source Robot Operating System}.
\newblock In: Proc. ICRA Workshop Open Source Software (2009).
\newblock
  \urlprefix\url{http://pub1.willowgarage.com/~konolige/cs225B/docs/quigley-icra2009-ros.pdf}

\bibitem{Remazeilles2006}
Remazeilles, A., Chaumette, F., Gros, P.: {3D navigation based on a visual
  memory}.
\newblock In: Proc. IEEE Int. Conf. Robot. and Autom. (ICRA), pp. 2719--2725
  (2006).
\newblock
  \urlprefix\url{http://ieeexplore.ieee.org/xpls/abs\_all.jsp?arnumber=1642112}

\bibitem{Royer2007}
Royer, E., Lhuillier, M., Dhome, M., Lavest, J.M.: {Monocular vision for mobile
  robot localization and autonomous navigation}.
\newblock Int. J. Comput. Vision (IJCV) \textbf{74}(3), 237--260 (2007).
\newblock
  \urlprefix\url{http://link.springer.com/article/10.1007/s11263-006-0023-y}

\bibitem{Simhon1998}
Simhon, S., Dudek, G.: {A global topological map formed by local metric maps}.
\newblock In: Proc. IEEE/RSJ Int. Conf. Intelligent Robot. and Syst. (IROS),
  pp. 1708--1714 (1998).
\newblock
  \urlprefix\url{http://ieeexplore.ieee.org/xpls/abs\_all.jsp?arnumber=724844}

\bibitem{Zhang2009}
Zhang, A.M., Kleeman, L.: {Robust appearance based visual route following for
  navigation in large-scale outdoor environments}.
\newblock Int. J. Robot. Research (IJRR) \textbf{28}(3), 331--356 (2009).
\newblock \urlprefix\url{http://ijr.sagepub.com/content/28/3/331.short}

\bibitem{Zhang2012}
Zhang, J., Singh, S., Kantor, G.: {Robust Monocular Visual Odometry for a
  Ground Vehicle in Undulating Terrain}.
\newblock In: Proc. Field and Service Robot. (FSR), pp. 311--326 (2012).
\newblock \urlprefix\url{http://link.springer.com/10.1007/978-3-642-40686-7
  http://dl.acm.org/citation.cfm?id=2407992
  http://link.springer.com/chapter/10.1007/978-3-642-40686-7\_21}

\bibitem{Zhao2010}
Zhao, L., Huang, S., Yan, L., Jianguo, J., Hu, G., Dissanayake, G.:
  {Large-Scale Monocular SLAM by Local Bundle Adjustment and Map Joining}.
\newblock In: Proc. IEEE Int. Conf. Control, Autom., Robot. and Vision
  (ICARCV), pp. 431--436 (2010).
\newblock
  \urlprefix\url{http://epress.lib.uts.edu.au/research/handle/10453/16326}

\end{thebibliography}

\end{document}